# A Comparative Study on Application of Class-Imbalance Learning for Severity Prediction of Adverse Events Following Immunization

Ning Chen, Zhengke Sun, and Tong Jia*, *Member, IEEE*

*Abstract*— **In collaboration with the Liaoning CDC, China, we propose a prediction system to predict the subsequent hospitalization of children with adverse reactions based on data on adverse events following immunization. We extracted multiple features from the data, and selected "hospitalization or not" as the target for classification. Since the data are imbalanced, we used various class-imbalance learning methods for training and improved the RUSBoost algorithm. Experimental results show that the improved RUSBoost has the highest Area Under the ROC Curve on the target among these algorithms. Additionally, we compared these class-imbalance learning methods with some common machine learning algorithms. We combined the improved RUSBoost with dynamic web resource development techniques to build an evaluation system with information entry and vaccination response prediction capabilities for relevant medical practitioners.**

## I. Background

As one of the basic national public health services, vaccination can effectively control infectious diseases. However, mishandling of adverse events following immunization (AEFI) may reduce public confidence in vaccine safety and adversely affect the implementation of vaccination to some extent. The number of AEFI reports from China has been on the rise for many years. In 2017, the total number of AEFIs in China was 210,181, of which 711 cases were severe abnormal reactions [1]. In the actual work of vaccination, it has become common for doctors in vaccination units to encounter AEFI. A system for predicting the severity of AEFI can help doctors deal with related events promptly and maintain public trust in vaccination. This study aimed to analyze and compare multiple machine learning algorithms and develop an AEFI assessment system using the best-performing machine learning algorithm among them. We used data on childhood AEFIs from 2009 to 2014 in Shenyang, China, for the study. Since the number of training samples varies greatly across classes. Therefore, we used multiple class-imbalance learning methods for training.

## II. Related Works

There is a background of initial applications of machine learning in dealing with clinical problems similar to vaccination. Machine learning has been used for the prediction of heart disease [2]. Neetu et al. showed that there are disparities in the performance of different machine learning algorithms on the same clinical dataset [3].

In addition, many algorithms have been proposed for the imbalanced classification problem. Liu et al. proposed two integrated learning methods based on undersampling: EasyEnsemble and BalanceCascade [4]. Seiffert et al. proposed RUSBoost to alleviate class imbalance [5].

There are already many user-friendly decision systems being developed. El-Ganainy et al. used machine learning to develop a support system capable of real-time clinical decision making [6]. Zhou et al. developed a user-friendly software platform for automatic radiomics modeling and analysis [7]. To get the optimal algorithm, users can compare multiple algorithms by AUC, accuracy, precision, and other indicators.

## III. Machine Learning Algorithms

### A. EasyEnsemble

EasyEnsemble is an algorithm for solving the problem of useful information loss during undersampling. The algorithm first uses an undersampling method to randomly select a subset of the same size as the minority class from the majority class and then uses the AdaBoost algorithm to learn that subset with the minority class set to obtain a classifier. The authors of this algorithm experimentally show that the algorithm performs better than many existing imbalance classification problems on some common imbalance classifier metrics [4].

### B. Balanced Random Forest

The Balanced Random Forest (BRF) algorithm is different from the simple superposition of undersampling and random forest, but it improves the sampling strategy of the random forest algorithm. The algorithm is shown below [8].

- For the sample set X, Bootstrap sampling is first performed from the small class to obtain K samples, and then sampling is performed from the large class to obtain K samples as well.

- For each sample with M attributes, when each node of the CART tree needs to split, m attributes are randomly selected, and then some strategy (e.g., information gain) is used from these m attributes to select 1 attribute as the splitting attribute for that node.

- The size of the CART tree should be as large as possible and not be cropped.

Repeat the above steps to build M trees, and make the final voting decision.

All authors are with the College Of Information Science and Engineering, Northeastern University, Shenyang, China.

*Contacting Author: Tong Jia is with the College Of Information Science and Engineering, Northeastern University, Shenyang, China. (phone: +86 13332421509; email: jiatong@ise.neu.edu.cn).

## C. RUSBoost and Our Improvement

The RUSBoost algorithm is similar to the SMOTEBoost algorithm. the SMOTEBoost algorithm solves the class imbalance problem by performing SMOTE oversampling of samples in each iteration of AdaBoost.M2, while the RUSBoost algorithm solves the problem of the high computational complexity of the SMOTEBoost algorithm by replacing the SMOTE oversampling with the random under-sampling method to obtain a simpler and faster model [5].

RUSBoost initially assigns the same weights to all samples and increases the weights of misclassified samples in the process. While in the SVM algorithm, the support vector is the key to classification. When using both algorithms for classification, some samples are more important than others, and it is these samples that influence the classification. By analogy, we conjecture that those samples that mainly affect the classification result are highly overlapping when both algorithms are tackling the same classification problem. Correctly classifying the samples that show importance in one algorithm is likely to improve the performance of the other algorithm. Therefore, we first pre-classify the samples using SVC, select the support vectors, and increase their weights appropriately. The changed weights are used as the initial weights of RUSBoost and let RUSBoost focus on training these samples to get a better performing model.

## D. Performance Metrics

We use the following evaluation metrics, where F1, G-mean, and Area Under the Receiver Operating Characteristic Curve (AUC) [9] values are commonly used to evaluate unbalanced classification problems.

TABLE I. CONFUSION MATRIX

|  | Predicted Positive Class | Predicted Negative Class |
|---|---|---|
| Actual Positive Class | TP (True Positives) | FN (False Negatives) |
| Actual Negative Class | FP (False Positives) | TN (True Negatives) |

$$\text{True Positive Rate}(Acc^+) = \frac{TP}{TP+FN}$$
$$\text{True Negtive Rate}(Acc^-) = \frac{TN}{TN+FP}$$
$$Precision = \frac{TP}{TP+FP}$$
$$Recall = \frac{TP}{TP+FN} = Acc^+$$
$$Specificity = \frac{TN}{TN+FP} = Acc^-$$
$$Accuracy = \frac{TN+TP}{TN+TP+FN+FP}$$
$$G\text{-}mean = \sqrt{Acc^- \times Acc^+}$$
$$F_1 = \frac{2 \times Precision \times Recall}{Precision + Recall}.$$
(1)

## IV. EXPERIMENT AND RESULT

In this section, we trained and evaluated some common machine learning algorithms, and then performed hyperparameter optimization for several integrated learning algorithms that cope with the imbalance problem. Grid search is often used in hyperparameter optimization problems but Bergstra et al. pointed out that random search is more effective than grid search in the optimization of hyperparameters [10]. We decided to combine both approaches, for parameters with small value space we chose to use grid search, and for parameters with larger value space we used random search. Then their performance in this problem is compared quantitatively and the optimal algorithm is selected for subsequent use.

### A. Data Pre-processing

We used child vaccination data from 2009-2014 in Shenyang, China, for our study, and we selected 1315 data from the dataset. First, we performed the feature selection (Fig. 1). The missing data were processed by padding. When processing the age of vaccination, some sample values were found to be too large, so these records were removed. The prediction of whether a patient is subsequently hospitalized or not is a binary problem.

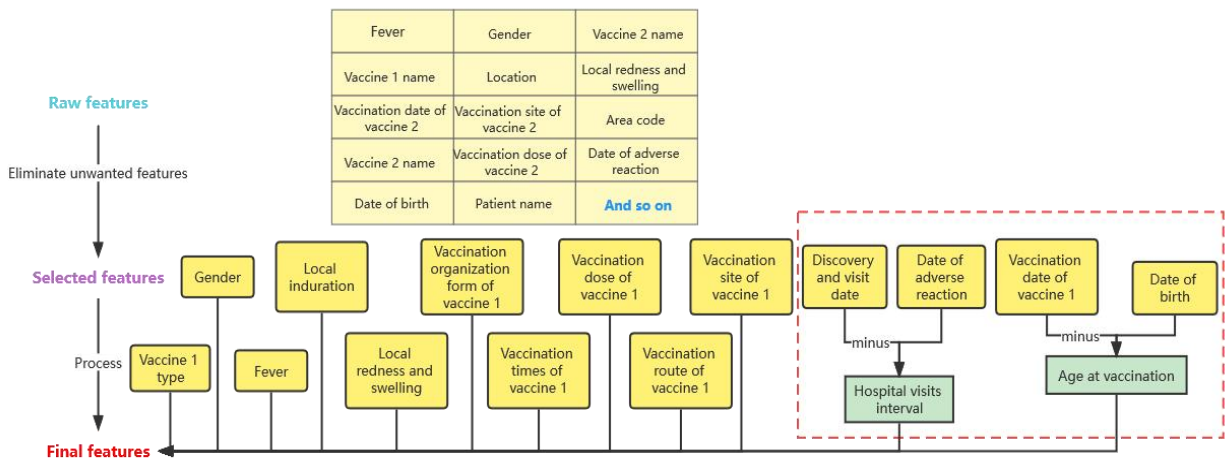

Figure 1. Feature selection.

TABLE II. EVALUATION METRICS OF SOME COMMON ALGORITHMS

| Algorithm | Accuracy | Precision | Recall | $F_1$ | AUC | Confusion Matrix |
|---|---|---|---|---|---|---|
| Random Forest | 0.958 | 0.973 | 0.984 | 0.978 | 0.723 | [[364 6] [ 10 3]] |
| SVC (linear kernel) | 0.966 | 0.966 | 1 | 0.983 | 0.585 | [[370 0] [ 13 0]] |
| Decision Tree | 0.948 | 0.973 | 0.973 | 0.973 | 0.581 | [[360 10] [ 10 3]] |
| Logistic Regression | 0.969 | 0.971 | 0.997 | 0.984 | 0.557 | [[369 1] [ 11 2]] |
| KNeighbors | 0.963 | 0.966 | 0.997 | 0.981 | 0.519 | [[369 1] [ 13 0]] |
| LightGBM | 0.948 | 0.97 | 0.976 | 0.973 | 0.588 | [[361 9] [ 11 2]] |
| XGBoost | 0.963 | 0.973 | 0.989 | 0.981 | 0.575 | [[366 4] [ 10 3]] |
| AdaBoost | 0.958 | 0.968 | 0.989 | 0.979 | 0.559 | [[366 4] [ 12 1]] |

We split the original 1315 data population into training and testing samples. Finally, we encoded the input and the output.

## B. Experiment Results of Some common Machine Learning Algorithms

We first use some common machine learning algorithms for training, and the training results are shown in Table II. It can be seen that some trained models have high values of accuracy, recall, and precision, but these algorithms tend to classify samples into the major class. However, in this study, the misclassification cost of the minority class is high, which indicates that these algorithms perform poorly in this problem. Moreover, indicators such as accuracy, recall, and precision are difficult to accurately evaluate the model used for this problem.

We will take three integrated learning methods dedicated to the imbalanced classification problem below. AUC values are commonly used to evaluate class imbalance problems [11][12][13]. So we use AUC as the evaluation metric for the objective of parameter search.

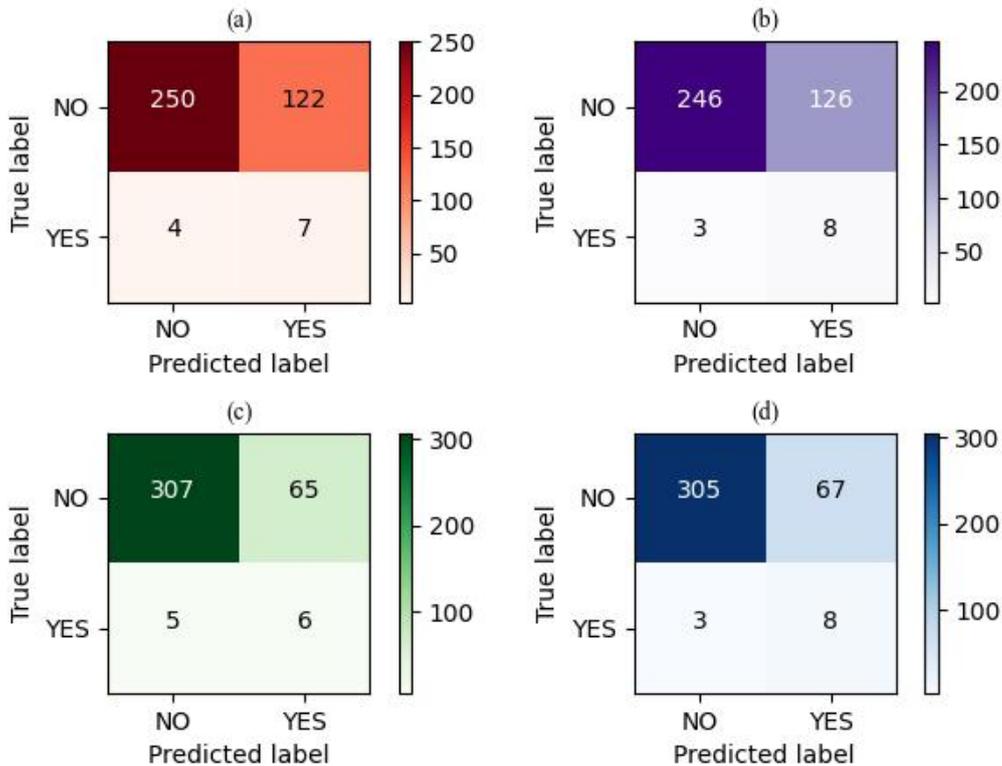

Figure 2. The confusion matrix of EE (a), BRF (b), RUSBoost (c), RUSBoost with SVC (d).

## C. Experiment Results of class-imbalance learning methods

The experimental results are shown in Fig. 2, Table III. In these three unimproved class-imbalanced learning algorithms, EasyEnsemble performs the worst, having the lowest AUC. Its training time is at least five times longer than the other two algorithms. Compared to the other two algorithms, EasyEnsemble is not suitable for solving this problem. Both EasyEnsemble and RUSBoost use the AdaBoost algorithm, but RUSBoost has a higher AUC value and it runs more efficiently. BRF has the highest AUC value. The overall performance of the RUSBoost and BRF is similar.

To get the improved RUSBoost model, first, we trained the polynomial kernel SVC to obtain the support vectors. Then we calculated the weights of each sample in RUSBoost, and output the vectors with the same number of support vectors as those in SVC, which are called "support vectors" in RUSBoost. We conducted ten iterations of the experiment and calculated that the average percentage of support vectors that are the same is 83.9%. This initially validates that, when tackling this problem, the two algorithms are highly overlapping in the samples which mainly affect the classification results.

It is clear that increasing the weights of those samples that are support vectors in SVC leads to an improvement in the classification results of RUSBoost. The performance metrics of the improved RUSBoost outperform the three class-imbalanced algorithms mentioned above. It can be demonstrated that in this problem, increasing the weights of the support vectors obtained in SVC can improve the classification effect of RUSBoost.

TABLE III. COMPARISON OF THREE IMBALANCE CLASSIFICATION ALGORITHMS

| Algorithm | Easy-Ensemble | BRF | RUSBoost | RUSBoost with SVC |
|---|---|---|---|---|
| Precision | 0.98 | **0.99** | 0.98 | **0.99** |
| Recall | 0.67 | 0.66 | **0.83** | 0.82 |
| Specificity | 0.64 | **0.73** | 0.55 | **0.73** |
| F₁ | 0.80 | 0.79 | **0.90** | **0.90** |
| G-mean | 0.65 | 0.69 | 0.67 | **0.77** |
| AUC | 0.73 | **0.83** | 0.82 | **0.83** |

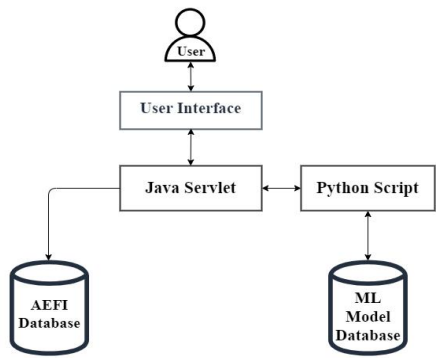

Figure 3. Architecture of the web application.

Figure 4. The user interface.

## V. APPLICATIONS

Due to the excellent performance of the improved RUSBoost, we finally developed a web application based on it. The overall framework is shown in Fig. 3. Users interact with the user interface (Fig. 4) using HyperText Markup Language and Cascading Style Sheets technology. When information entry is performed, the servlet accepts the data and stores the added data directly into the database. The data in the database can be used for subsequent updates of the model. For inoculation response prediction, the form submitted by the user is received by a java servlet. the servlet will call a python script that uses the machine learning model to get the prediction results ready to be presented to the user. Finally, JavaScript is used to dynamically display or clear results.

## VI. CONCLUSION

We first used ordinary machine learning methods for training by analyzing data on AEFI, but the results were not good. It also illustrated that some metrics are not suitable for evaluating imbalanced classification problems, and different evaluation metrics need to be selected for different problems to evaluate machine learning models. Given the imbalance of the sample, we used a variety of class-imbalance learning methods. Finally, we obtained a model with an AUC of 0.83 by improving the RUSBoost. The experimental results verify that those samples that mainly affect the classification results are highly overlapping between SVC and RUSBoost when tackling this problem. Correctly classifying the samples that show importance among SVC does improve the performance of RUSBoost.

A simple prediction system based on this model was developed for the subsequent hospitalization of those children who developed adverse reactions to vaccination. Hospitalization after the occurrence of adverse vaccination reactions can indicate the severity of AEFI to some extent. This system can help physicians to make a simple analysis of the severity of AEFI.

In future work, we plan to further test and extend our conjectures and look for additional indicators to analyze the severity of AEFI comprehensively. We will also optimize the interface and architecture of the software to make it more suitable for clinical use and to help the vaccination program continue to progress steadily.